\documentclass[11pt]{article}

\usepackage{acl}

\usepackage{times}
\usepackage{latexsym}
\usepackage{amsmath}
\usepackage{amsfonts}
\usepackage{booktabs}
\usepackage{multirow}
\usepackage{minted}
\usepackage{array}
\usepackage{multicol}
\usepackage{booktabs}
\usepackage{makecell}
\usepackage{graphicx}  % for \includegraphics
\usepackage{subcaption}
\usepackage{tabularx}
\usepackage[T1]{fontenc}
\usepackage[utf8]{inputenc}

\usepackage{microtype}

\usepackage{inconsolata}

\usepackage{graphicx}
\usepackage{tikz}
\usepackage{listings}
\usepackage{graphicx}

\usepackage[table]{xcolor} 
\usepackage{booktabs}
\usepackage{multirow}

\definecolor{colorCoT}{HTML}{E1F5FE}
\definecolor{colorFewShot}{HTML}{FFF9C4}
\definecolor{colorLoRA}{HTML}{E8F5E9}
\definecolor{colorDPO}{HTML}{FCE4EC}
\definecolor{colorAligner}{HTML}{F3E5F5}

\usepackage{tcolorbox}   % For prompt and hyperparameter boxes
\usepackage{verbatim}   % For verbatim environments
\usepackage{fancyvrb}   % For Verbatim with line breaks
\usepackage{amsmath}    % For math environments (if used elsewhere)
\usepackage{xcolor}     % For colored boxes

\title{YNTP-100: A Benchmark for Your Next Token Prediction with 100 People}

\author{Shiyao Ding \\
  Kyoto University \\
  Kyoto, Japan \\
  \texttt{ding@i.kyoto-u.ac.jp} \\\And
  Takayuki Ito \\
  Kyoto University \\
  Kyoto, Japan \\
  \texttt{ito@i.kyoto-u.ac.jp} \\}

\begin{document}
\maketitle
\begin{abstract}
Large language models (LLMs) trained for general \textit{next-token prediction} often fail to generate responses that reflect how specific individuals communicate. Progress on personalized alignment is further limited by the difficulty of collecting real-world personal communication data due to privacy constraints.
We propose Your Next Token Prediction (YNTP), a task that formulates personalized response generation as token-level prediction conditioned on user interaction history. We introduce \textbf{YNTP-100}, a benchmark built from multilingual multi-day human--agent conversations with 100 people, enabling systematic evaluation of user-specific response behavior. We evaluate external (parameter-preserving) and internal (parameter-updating) alignment methods using metrics of substance similarity and stylistic consistency. The dataset and results are publicly available at: \url{https://github.com/AnonymousHub4Submissions/YNTP100}.
\end{abstract}

\section{Introduction}
Large language models (LLMs) are trained to optimize \textit{next-token prediction}, yet they often fail to generate responses that reflect how a specific individual communicates in everyday settings such as emails or social messages. While recent work on \textit{personalized alignment} aims to adapt LLMs to individual users, existing formulations do not adequately model how people respond during interpersonal communication.

Most existing personalization benchmarks, including LaMP/LongLaMP~\cite{salemi2023lamp,kumar2024longlamp}, P-SOUPS~\cite{jang2023personalized}, and PRISM~\cite{kirk2024prism}, conceptualize personalization primarily as \textit{task-level customization}, where models adapt surface properties such as writing style or content selection (e.g., title generation or abstract writing).
This framing abstracts away the sequential and interactive nature of real-world communication—such as replying to emails or social media messages—where responses are shaped by an individual’s habitual tone and communicative stance over time.
As a result, current benchmarks are limited in evaluating whether models can generate responses in the way a specific person would.

Enabling such personalized response generation faces two key challenges: (1) authentic SNS or email histories are difficult to obtain due to privacy and ethical constraints, and (2) most existing benchmarks are English-centric, limiting their generality across languages and cultures.

To address these challenges, we propose Your Next-Token Prediction (YNTP), a new task that formulates personalized alignment as a fine-grained prediction problem: given a conversational context, the objective is to predict how an individual user would respond. YNTP-100 is constructed from multilingual human--agent conversations involving 100 participants, with over 30 users each in English, Japanese, and Chinese. Participants engage in five-day dialogue sessions with LLM-driven non-player characters (NPCs) in a controlled interaction environment designed to elicit natural daily communication.

The NPCs are psychologically grounded and follow structured finite-state machines (FSMs). This design enables targeted yet natural exploration of user traits, such as extraversion or intuition, while maintaining consistency across interactions. By observing users across multiple days and conversational contexts, the goal of YNTP is to capture each individual’s internal response patterns,including linguistic choices, emotional tendencies, and decision-making behavior, as they evolve over time. The benchmark task is to predict a user’s response on the final day given their preceding interaction history, providing a testbed for evaluating a model’s ability to capture continuity, adaptation, and personal consistency.

We evaluate both external (parameter-preserving) and internal (parameter-updating) alignment methods on YNTP, establishing the first quantitative baseline for personalized response generation under this task.
For evaluation, rather than adopting the general alignment paradigm of 3H (Helpful, Harmless, Honest), we introduce a 2S principle that characterizes personalized responses along two essential dimensions: Substance (what to say) and Style (how to say it).
Overall, YNTP represents a step toward the “last mile” of alignment, moving LLMs from generic communicators toward more consistent and user-aligned interaction behavior.

\section{Related Work}
\subsection{Personalized Alignment Datasets}
The most widely used personalized benchmarks are the LaMP family~\cite{salemi2023lamp}, which unify multiple personalization tasks such as title generation and email rewriting conditioned on user histories. LongLaMP~\cite{kumar2024longlamp} extends this framework to long-form generation, including reviews and blog posts. These benchmarks standardize evaluation and emphasize the role of retrieval and user history, but primarily model how users adapt content rather than how they engage in interpersonal communication.  
Beyond LaMP, P-SOUPS~\cite{jang2023personalized} and PRISM~\cite{kirk2024prism} explore personalization across multiple domains and dialogue settings using user profiles or stylistic metadata. Recent datasets such as PersonalLLM~\cite{zollo2024personalllm} and PERSONA~\cite{castricato2024persona} further investigate explicit user representations for persona-consistent generation. Overall, these datasets focus on modeling user-specific writing styles and behavioral patterns in task-oriented settings.

Another line of work studies alignment through preference signals rather than explicit user identities. Datasets such as FLASK~\cite{ye2023flask}, REGEN~\cite{sayana2025beyond}, ALOE~\cite{wu2024aligning}, and PREFEVAL~\cite{zhao2025llms} collect human or model feedback to guide alignment, often in the context of RLHF or related paradigms. While effective for improving general alignment quality, these datasets target population-level preferences rather than personalized response modeling. A summary of these datasets is provided in Appendix~A (Table~\ref{tab:personalization-datasets}).

\subsection{Personalized Alignment Methods}
Existing approaches can be broadly categorized into two paradigms: external (parameter-preserving) and internal (parameter-updating).

\textbf{External alignment methods} keep the LLM frozen and incorporate user information at inference time. Prompt-based approaches inject user histories or summaries directly into the input \cite{christakopoulou2023large, richardson2023integrating, tang2024step}. Retrieval-augmented generation (RAG) extends this idea by retrieving user-specific documents or prior interactions as in-context examples \cite{salemi2023lamp}, sometimes with optimized retrieval strategies \cite{salemi2024optimization}. Other approaches represent users through embeddings or structured profiles for persona-conditioned inference \cite{liu2024llms+}. More recent techniques steer decoding or latent activations to reflect user traits, enabling scalable personalization without retraining or direct access to private data.

\textbf{Internal alignment methods} directly adapt model parameters to capture user-specific behavior. Parameter-efficient fine-tuning (PEFT) techniques, such as LoRA and prefix-tuning \cite{hu2022lora, finedemocratizing}, update a small subset of weights to reduce computational cost. Personalized RLHF (P-RLHF) \cite{li2024personalized} further aligns models using user-level feedback, enabling fine-grained personalization at the expense of additional data collection. 
In a complementary direction, Aligner~\cite{ji2024aligner} applies a post-hoc alignment module that rewrites an initial model output to improve consistency with user-specific linguistic and stylistic patterns, without modifying the underlying model parameters.
Recent inference-stage variants, including PAD \cite{chen2024pad} and CHAMELEON \cite{nguyen2025yo}, steer token probabilities or latent activations using personalized rewards, offering lightweight yet effective adaptation.

\section{YNTP-100 Benchmark}

\subsection{Task Formulation}
We define YNTP) as a personalized response generation task: given an incoming message $x$, the objective is to predict how a specific user $u$ would respond at the token level.

Let $y^u = (y^u_1, \dots, y^u_m)$ denote the response written by user $u$. The model generates a personalized response $\hat{y}^u$ conditioned on the input message and user context $U^u$, which consists of optional profile information $p^u$ and interaction history $H^u$:
\[
U^u = (p^u, H^u).
\]
The user-conditioned generation objective follows an autoregressive formulation:
\[
P_\theta(\hat{y}^u \mid x, U^u)
= \prod_{t=1}^{m} P_\theta(\hat{y}^u_t \mid \hat{y}^u_{<t}, x, U^u),
\]
where $\theta$ denotes model parameters. Compared to standard next-token prediction, YNTP explicitly conditions generation on user-specific context, enabling evaluation of personalized response behavior.

\paragraph{Learning Objective.}
Given a dataset of user-specific dialogue instances
\[
\mathcal{D} = \{(x_i, y^u_i, U^u_i)\}_{i=1}^{N},
\]
models are trained to minimize the token-level negative log-likelihood:
\[
\min_{\theta} \;
- \mathbb{E}_{(x, y^u, U^u) \sim \mathcal{D}}
\left[
\sum_{t=1}^{|y^u|}
\log P_\theta(y^u_t \mid y^u_{<t}, x, U^u)
\right].
\]
This objective evaluates whether a model can reproduce user-specific token distributions rather than generic responses.

\subsection{Human--Agent Dialogue System}
To construct the YNTP-100 benchmark, we collect multi-day conversational data using an LLM-driven human--agent dialogue system. Users interact with multiple non-player characters (NPCs) in a shared conversational environment, producing paired message--response data $(x, y^u)$ together with contextual metadata $U^u$.

\begin{figure}[t]
  \centering
  \includegraphics[width=0.9\linewidth]{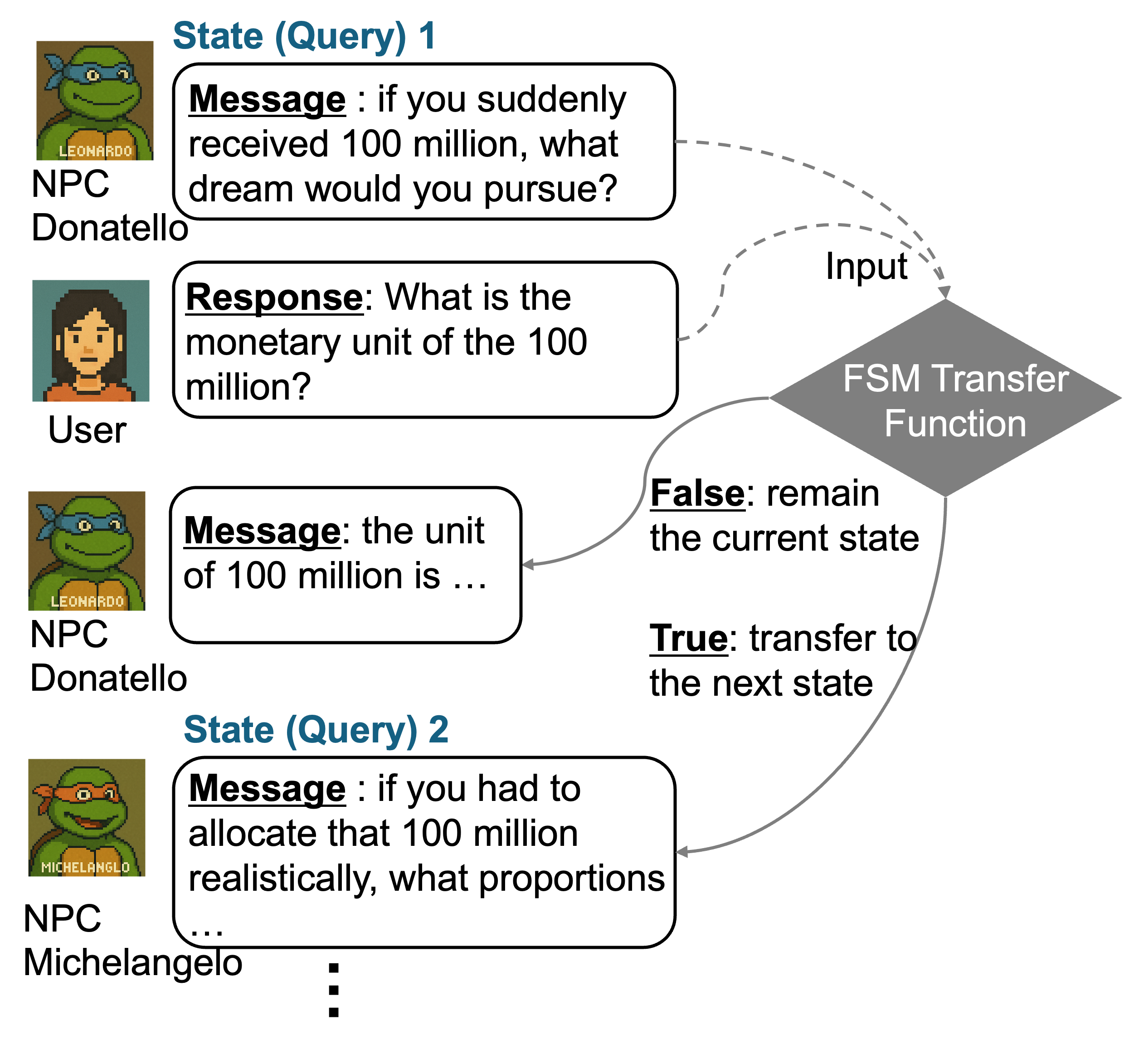}
  \caption{FSM-based control flow of the LLM-driven multi-NPC dialogue system. Each state corresponds to a predefined conversational query. User responses are evaluated by a lightweight check function that determines whether to transition to the next state or remain in the current one.}
  \label{fig:fsm}
\end{figure}

Dialogue progression is governed by a lightweight finite-state machine (FSM) that structures interaction flow across multiple days. As illustrated in Figure~\ref{fig:fsm}, each FSM state corresponds to a predefined conversational query posed by an NPC. For example, in State (Query) 1, the NPC asks an open-ended question (e.g., about how the user would use a large sum of money). If the user’s response is incomplete or seeks clarification, the FSM’s transfer function keeps the dialogue in the current state and triggers a follow-up prompt. Once the response satisfies basic relevance or sufficiency criteria, the dialogue transitions to State (Query) 2, where a different NPC poses a related but more specific question. This mechanism allows the system to elicit comparable responses under controlled conditions while preserving natural, multi-turn conversational flow.

\begin{figure}[ht]
  \centering
  \includegraphics[width=1\linewidth]{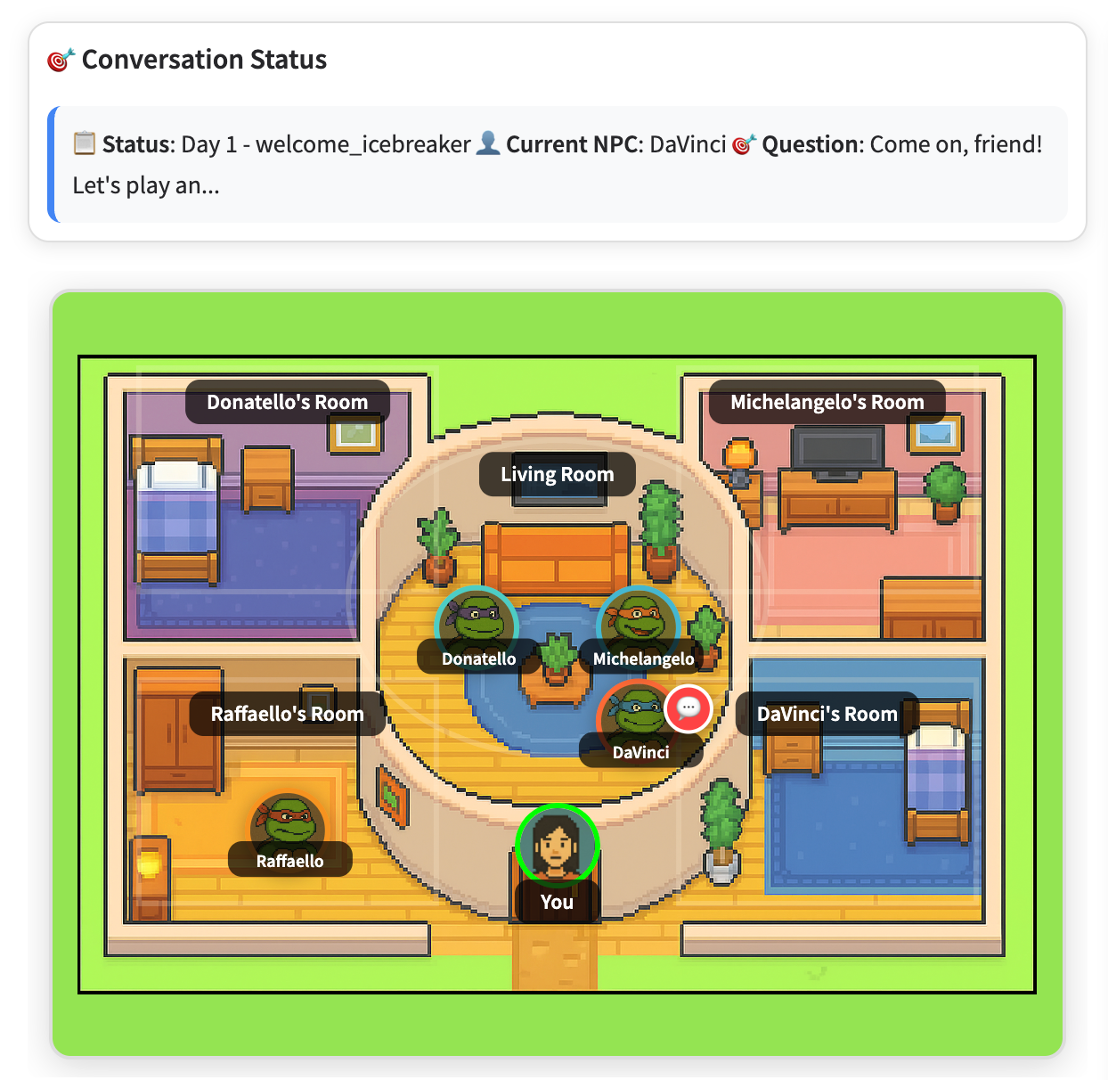}
  \caption{Dialogue interface used for data collection. The shared-house environment and status panel visualize user--NPC interactions across multiple dialogue threads and days.}
  \label{fig:framework}
\end{figure}

Figure~\ref{fig:framework} illustrates the dialogue interface used during data collection. The interface presents a shared-house environment and a status panel that visualize ongoing user--NPC interactions across multiple dialogue threads. Its purpose is to facilitate consistent and engaging interaction during multi-day sessions, rather than to impose specific behavioral constraints.

To introduce structured variation in dialogue prompts, FSM states are annotated with coarse-grained personality-related labels inspired by the MBTI (Myers–Briggs Type Indicator) taxonomy. These labels are used solely as an interpretable scheme for organizing queries and ensuring coverage of diverse conversational situations. They do not assume psychological validity, fixed user traits, or explicit personality classification. FSM transitions depend only on response relevance or sufficiency, ensuring consistent data collection across users.

\subsection{Data Collection and Benchmark Construction}
Each interaction yields a dialogue instance $(x, y^u)$, where $x$ denotes the NPC’s message and $y^u$ the response from user $u$, annotated with metadata including the day index, NPC identity, and FSM state. All conversations are stored in a structured JSON format as ordered message--response pairs.

For benchmark construction, we use interactions from the first four days (\textbf{training data}) of each user to model user-specific behavior, and reserve the fifth day (\textbf{test data}) for evaluation. During testing, models generate responses to fifth-day messages, which are compared against the corresponding user responses to assess personalized alignment.

The benchmark comprises three multilingual datasets collected under identical protocols: 34 English users, 33 Chinese users, and 33 Japanese users. This design supports evaluation across languages while controlling for interaction structure and data availability. Further details of participant procedures, recruitment, compensation, and data consent are provided in Appendix~B.

\section{EXPERIMENTS SETUP}

We evaluated based on the colloected 100 users dataset to evaluate various personlized alignment methonds and LLMs effectiveness.

\subsection{Baseline Methods}
We compare representative strategies for personalized alignment, covering both external (parameter-preserving) and internal (parameter-updating) alignment methods:

\textbf{Prompt Engineering (Zero-shot)} \cite{radford2019language} conditions generation on a brief user description (e.g., profile cues) without providing any historical examples, serving as a minimal personalization baseline.

\textbf{Prompt Engineering (Few-shot)} \cite{brown2020language} augments the prompt with demonstration pairs from the user’s prior interactions, enabling the model to ground its generation in user-specific patterns observed from past responses.

\textbf{Chain-of-Thought (CoT)} \cite{wei2022chain} encourages the model to perform explicit intermediate reasoning before producing the final response, providing a reasoning-based prompting baseline for personalized generation.

\textbf{Fine-Tuning}: SFT/ PEFT (LoRA) \cite{ouyang2022training,hu2022lora} evaluates parameter-updating personalization under two practical settings. For \emph{closed-source} models (e.g., GPT-style APIs), we apply SFT when full-model fine-tuning is supported. For \emph{open-source} models (e.g., Llama and Qwen families), we apply PEFT (LoRA), which updates a small set of low-rank adapters using user-specific data.

\textbf{Direct Preference Optimization (DPO)} \cite{rafailov2023direct} trains models to prefer personalized outputs over generic ones using a preference-based objective defined over pairs of candidate responses.

\textbf{Aligner} \cite{ji2024aligner} applies a post-hoc alignment module that rewrites an initial model output to improve consistency with user-specific linguistic and stylistic patterns.

For reproducibility, the exact prompts and hyperparameters used for each method are reported in Appendix~E.

\subsection{Language Models}
We evaluate both closed-source and open-source language models.
\textbf{Closed-source Models} include gpt-3.5-turbo~\cite{ye2023comprehensive}, gpt-4o-mini~\cite{hurst2024gpt}, gemini-2.5-flash~\cite{comanici2025gemini}, and claude-sonnet-4-0~\cite{koyun2025evaluation}, which offer strong instruction-following and reasoning performance.
\textbf{Open-source Models} include DeepSeek-R1-Distill-Qwen-14B~\cite{guo2025deepseek}, Qwen3-14B~\cite{yang2025qwen3}, and Llama-3.1-8B-Instruct~\cite{grattafiori2024llama}. For Japanese-specific evaluation, we additionally include Llama-3-ELYZA-JP-8B~\cite{elyzallama2024}.

\subsection{Evaluation Metrics}
Unlike the 3H principle used in general alignment, for the YNTP task we propose a \textbf{2S Principle}, which characterizes personalized response generation along two essential dimensions:
 \emph{Substance} (\emph{what to say}) and \emph{Style} (\emph{how to say}).  
This principle reflects the core requirement of personalized alignment: a model should generate responses that are semantically appropriate to the input message while expressing them in a manner consistent with how a specific user typically communicates.

In practice, each dimension can be measured using a variety of metrics. In this paper, we select three representative metrics for each dimension. The complete set of six metrics used in our evaluation is summarized in Table~\ref{tab:personalization-metrics}.
 Detailed formulations of each metric are provided in Appendix~C.
 
\begin{table}[t]
\centering
\caption{Evaluation metrics for personalized response generation.}
\label{tab:personalization-metrics}
\small
\setlength{\tabcolsep}{4pt}
\begin{tabular}{p{2.0cm} p{5cm}}
\toprule
\textbf{Group \& Metrics} & \textbf{Description} \\
\midrule

\textbf{Substance}\\[-1mm]
\rule{2.2cm}{0.2pt} \\
M1: Word Mover’s Distance &
Measures semantic dissimilarity between predicted and reference responses using the minimal transport distance between word embeddings. Lower values indicate closer semantic alignment. \\
M2: Sentence Similarity &
Cosine similarity between sentence embeddings capturing overall semantic closeness at the sentence level. \\
M3: BLEU (Content Similarity) &
n-gram precision with a brevity penalty evaluating surface-level content overlap. \\

\midrule

\textbf{Style}\\[-1mm]
\rule{2.2cm}{0.2pt} \\
M4: Normalized Length Similarity &
Ratio of the shorter to the longer response length, measuring verbosity alignment. \\
M5: Type--Token Ratio (TTR) &
Lexical richness measured as the proportion of unique tokens in the response. \\
M6: History Similarity &
Embedding-based similarity between the generated response and the user’s historical responses. \\

\bottomrule
\end{tabular}
\end{table}

\section{Results and Analysis}

\begin{table*}
\centering
\caption{
Z-score normalized results across three languages: macro-averaged scores over 33 English users, 34 Chinese users, and 33 Japanese users.
Each value represents the z-score: (value - mean) / standard deviation for each metric column.
For M1--M6, the bigger values indicate better performance.
\colorbox{red!30}{Best values}, \colorbox{orange!30}{second-best}, and \colorbox{yellow!30}{third-best} are highlighted based on original rankings.
}
\label{tab:main-all-zscore}
\small
\setlength{\tabcolsep}{1.5pt}
\setlength{\fboxsep}{1pt}
\renewcommand{\arraystretch}{0.95}
\resizebox{\textwidth}{!}{
\begin{tabular}{p{2cm} p{1.5cm} r r r r r r | r r r r r r | r r r r r r}
\toprule
\multirow{2}{*}{\textbf{Method}} & \multirow{2}{*}{\textbf{Base Model}} &
\multicolumn{6}{c|}{\textbf{English (33 users)}} &
\multicolumn{6}{c|}{\textbf{Chinese (34 users)}} &
\multicolumn{6}{c}{\textbf{Japanese (33 users)}} \\
\cmidrule(lr){3-8} \cmidrule(lr){9-14} \cmidrule(lr){15-20}
 & & M1 & M2 & M3 & M4 & M5 & M6 & M1 & M2 & M3 & M4 & M5 & M6 & M1 & M2 & M3 & M4 & M5 & M6 \\
\midrule
\multirow{8}{=}{\parbox{2cm}{\raggedright Prompt Eng. (zero-shot)}}
& Gpt-3.5-turbo & 0.303 & 0.627 & -0.127 & -0.048 & 0.329 & 0.473 & -0.713 & -0.577 & -0.359 & -0.267 & -0.015 & -1.389 & -1.904 & -1.046 & -0.866 & -0.350 & -0.208 & -1.881 \\
 & Gpt-4o-mini & 0.306 & 0.597 & -0.449 & -0.576 & 0.051 & 0.029 & 0.562 & 0.212 & -0.446 & -0.406 & 0.085 & 0.366 & 0.551 & -0.457 & -0.315 & -0.433 & -0.064 & 0.103 \\
 & Gemini-2.5-flash & 0.289 & 0.085 & -1.188 & -1.181 & -0.767 & 0.389 & -0.052 & -0.645 & -1.003 & -1.238 & -0.923 & -0.248 & 0.254 & -0.984 & -1.029 & -1.234 & -0.975 & -0.570 \\
 & Claude-sonnet-4-0 & \colorbox{yellow!30}{0.367} & -0.232 & -1.300 & -1.267 & -0.431 & 0.632 & -0.726 & -1.005 & -1.011 & -1.094 & -0.556 & -0.774 & -0.275 & -1.052 & -0.999 & -0.994 & -0.388 & -0.656 \\
 & DeepSeek-R1-Distill-Qwen-14B & \colorbox{red!30}{0.394} & -0.181 & -1.147 & -1.420 & -1.123 & \colorbox{red!30}{1.348} & -0.898 & -0.858 & -0.910 & -1.370 & -1.043 & -1.026 & -3.705 & -1.267 & -1.334 & -1.501 & -1.176 & -1.821 \\
 & Qwen3-14B & 0.313 & 0.095 & -1.143 & -1.142 & -0.858 & 0.231 & 0.405 & -0.603 & -0.811 & -1.029 & -0.558 & 0.020 & 0.402 & -0.977 & -0.954 & -1.098 & -1.006 & -0.591 \\
 & Llama-3.1-8B-Instruct & 0.339 & 0.173 & -0.831 & -1.055 & -1.170 & 0.289 & -0.268 & -0.778 & -0.772 & -0.946 & -1.305 & -0.604 & -0.934 & -1.248 & -0.944 & -0.921 & -1.395 & -1.296 \\
 & Llama-3-ELYZA-JP-8B & -0.266 & -0.795 & -0.679 & -0.178 & 0.211 & -1.286 & -2.836 & -2.521 & -1.173 & -0.997 & -1.300 & -3.043 & -1.062 & -1.151 & -0.735 & -0.542 & -0.656 & -1.329 \\
\hline
\multirow{8}{=}{\parbox{2cm}{\raggedright Prompt Eng. (few-shot)}}
& Gpt-3.5-turbo & 0.244 & 0.766 & 0.462 & 0.190 & 0.506 & 0.849 & 0.642 & 0.648 & 0.539 & 0.206 & 0.409 & 0.570 & 0.552 & 0.561 & 0.134 & 0.266 & 0.377 & 0.382 \\
 & Gpt-4o-mini & 0.261 & \colorbox{red!30}{0.972} & 0.263 & 0.118 & 0.510 & 0.426 & 0.789 & 0.711 & -0.037 & 0.135 & 0.479 & 0.515 & 0.804 & 0.646 & 1.004 & 0.166 & 0.387 & 0.610 \\
 & Gemini-2.5-flash & 0.007 & 0.390 & \colorbox{orange!30}{1.557} & \colorbox{yellow!30}{1.446} & \colorbox{yellow!30}{0.910} & 0.174 & 0.732 & 0.793 & \colorbox{yellow!30}{1.487} & 1.261 & 0.946 & 0.810 & 0.757 & 1.198 & 1.486 & 1.382 & 1.045 & 1.071 \\
 & Claude-sonnet-4-0 & 0.275 & 0.705 & 0.730 & 0.421 & 0.471 & 0.551 & \colorbox{red!30}{1.010} & 0.797 & 1.298 & 0.723 & 0.750 & 0.786 & \colorbox{red!30}{0.852} & 1.093 & \colorbox{yellow!30}{1.670} & 1.013 & 0.860 & 0.992 \\
 & DeepSeek-R1-Distill-Qwen-14B & 0.361 & -3.705 & -1.467 & -1.733 & -2.162 & 0.245 & -0.949 & -1.168 & -1.185 & -1.560 & -1.762 & -1.108 & -0.393 & -1.686 & -1.271 & -1.631 & -1.973 & -1.858 \\
 & Qwen3-14B & 0.270 & \colorbox{yellow!30}{0.801} & 1.246 & 0.696 & 0.507 & 0.447 & 0.712 & 0.725 & 0.884 & 0.617 & 0.618 & 0.588 & 0.651 & 0.680 & 0.393 & 0.716 & 0.552 & 0.635 \\
 & Llama-3.1-8B-Instruct & 0.074 & -0.384 & -0.941 & -1.055 & -2.726 & \colorbox{orange!30}{1.141} & 0.174 & -0.821 & -0.803 & -0.843 & -1.949 & -0.256 & -0.220 & -1.069 & -1.031 & -0.798 & -1.847 & -1.003 \\
 & Llama-3-ELYZA-JP-8B & -0.064 & 0.020 & \colorbox{yellow!30}{1.521} & 1.029 & 0.775 & -0.006 & 0.031 & 0.663 & 0.404 & 0.795 & 0.582 & 0.465 & -0.134 & 0.688 & 0.036 & 0.924 & 0.711 & 0.415 \\
\hline
\multirow{6}{=}{\parbox{2cm}{\raggedright Chain-of-Thought}}
& Gpt-3.5-turbo & 0.235 & 0.635 & 0.415 & 0.913 & 0.812 & 0.584 & 0.775 & \colorbox{red!30}{0.947} & 1.281 & 1.169 & 0.892 & \colorbox{red!30}{0.831} & 0.512 & 1.057 & 0.507 & 0.959 & 0.869 & 0.891 \\
 & Gpt-4o-mini & 0.275 & \colorbox{orange!30}{0.840} & 0.798 & 0.636 & 0.615 & 0.483 & 0.872 & \colorbox{orange!30}{0.944} & 0.659 & 0.739 & 0.752 & 0.765 & 0.779 & 1.107 & 1.219 & 0.884 & 0.792 & 1.034 \\
 & Gemini-2.5-flash & 0.026 & 0.497 & 1.106 & 1.290 & 0.862 & 0.170 & 0.797 & 0.889 & \colorbox{red!30}{1.970} & 1.364 & \colorbox{yellow!30}{1.032} & 0.819 & \colorbox{yellow!30}{0.810} & 1.157 & \colorbox{red!30}{2.011} & 1.353 & 1.034 & 1.106 \\
 & Claude-sonnet-4-0 & 0.207 & 0.782 & 1.069 & 1.157 & 0.720 & 0.608 & \colorbox{yellow!30}{0.945} & 0.822 & 1.320 & 1.192 & 0.898 & \colorbox{yellow!30}{0.822} & \colorbox{orange!30}{0.830} & 1.198 & \colorbox{orange!30}{1.747} & 1.199 & 1.016 & 1.153 \\
 & DeepSeek-R1-Distill-Qwen-14B & 0.170 & 0.467 & 0.179 & 0.508 & 0.534 & 0.290 & \colorbox{orange!30}{0.967} & 0.833 & 1.238 & 0.871 & 0.708 & 0.748 & 0.430 & 0.411 & 0.234 & 0.423 & 0.458 & 0.398 \\
 & Qwen3-14B & -0.116 & 0.091 & 0.979 & 1.254 & \colorbox{orange!30}{0.942} & -0.076 & 0.020 & 0.601 & -0.169 & 1.061 & \colorbox{red!30}{1.048} & 0.412 & -0.143 & 0.213 & -0.356 & 0.166 & -0.372 & 0.122 \\
\hline
\multirow{5}{=}{\parbox{2cm}{\raggedright Fine-tuning}}
& Gpt-3.5-turbo & -0.222 & -0.443 & 0.846 & 1.262 & 0.897 & -0.312 & 0.389 & 0.803 & 0.693 & \colorbox{yellow!30}{1.397} & \colorbox{orange!30}{1.042} & 0.710 & 0.243 & \colorbox{yellow!30}{1.223} & 0.214 & \colorbox{yellow!30}{1.492} & \colorbox{red!30}{1.587} & \colorbox{yellow!30}{1.158} \\
 & Gpt-4o-mini & 0.031 & -0.040 & \colorbox{red!30}{2.648} & \colorbox{orange!30}{1.497} & 0.700 & 0.957 & 0.653 & \colorbox{yellow!30}{0.923} & \colorbox{orange!30}{1.948} & \colorbox{red!30}{1.578} & 1.002 & 0.809 & 0.403 & \colorbox{orange!30}{1.257} & 1.421 & \colorbox{orange!30}{1.505} & \colorbox{yellow!30}{1.272} & \colorbox{orange!30}{1.212} \\
 & Qwen3-14B & -0.155 & -0.252 & 1.267 & 0.910 & 0.736 & 0.185 & 0.450 & 0.783 & 1.113 & 1.086 & 0.838 & 0.647 & 0.436 & 0.876 & 0.473 & 0.827 & 0.798 & 0.878 \\
 & Llama-3.1-8B-Instruct & -0.163 & -1.160 & -0.406 & 0.830 & 0.790 & -0.870 & -0.373 & 0.447 & -0.540 & 0.583 & 0.836 & 0.348 & -0.078 & 0.987 & -0.414 & 1.003 & 1.239 & 0.939 \\
 & Llama-3-ELYZA-JP-8B & 0.033 & -0.492 & 1.241 & \colorbox{red!30}{1.519} & 0.634 & \colorbox{yellow!30}{0.979} & 0.409 & 0.880 & 0.688 & \colorbox{orange!30}{1.399} & 0.925 & \colorbox{orange!30}{0.826} & 0.376 & \colorbox{red!30}{1.280} & 1.064 & \colorbox{red!30}{1.618} & \colorbox{orange!30}{1.319} & \colorbox{red!30}{1.302} \\
\hline
\multirow{3}{=}{\parbox{2cm}{\raggedright DPO}}
& Qwen3-14B & 0.309 & 0.081 & -0.908 & -1.173 & -0.884 & 0.306 & 0.326 & -0.735 & -0.891 & -1.081 & -0.592 & -0.096 & 0.536 & -1.055 & -0.966 & -1.093 & -0.934 & -0.612 \\
 & Llama-3.1-8B-Instruct & \colorbox{orange!30}{0.385} & -0.021 & -1.064 & -1.221 & -1.030 & 0.488 & 0.480 & -0.374 & -0.715 & -0.771 & -0.688 & 0.092 & 0.504 & -0.941 & -0.853 & -0.840 & -1.174 & -0.542 \\
 & Llama-3-ELYZA-JP-8B & -6.454 & -3.970 & -0.650 & -0.256 & \colorbox{red!30}{0.981} & -3.387 & -2.265 & -2.672 & -1.263 & -1.065 & -1.511 & -2.637 & 0.095 & -0.954 & -0.808 & -0.839 & -0.866 & -0.963 \\
\hline
\multirow{3}{=}{\parbox{2cm}{\raggedright Aligner}}
& Qwen3-14B & 0.276 & -1.561 & -1.423 & -1.622 & -2.532 & -3.494 & -2.387 & -1.460 & -1.208 & -1.475 & -1.963 & -1.645 & -2.502 & -1.406 & -1.320 & -1.412 & -1.753 & -1.784 \\
 & Llama-3.1-8B-Instruct & 0.225 & 0.078 & -1.488 & -1.622 & -2.722 & -1.216 & -1.791 & -1.201 & -1.204 & -1.507 & -2.612 & -1.354 & -1.310 & -1.343 & -1.308 & -1.474 & -2.363 & -1.441 \\
 & Llama-3-ELYZA-JP-8B & -0.850 & -0.639 & -0.565 & -0.589 & -0.195 & -2.647 & -2.722 & -2.464 & -1.272 & -1.159 & -1.408 & -2.669 & -2.593 & -1.554 & -1.120 & -1.128 & -0.902 & -1.859 \\
\bottomrule
\end{tabular}
}
\end{table*}

\subsection{Overall Analysis}
\begin{figure*}[htbp]
    \centering
    \includegraphics[width=\textwidth]{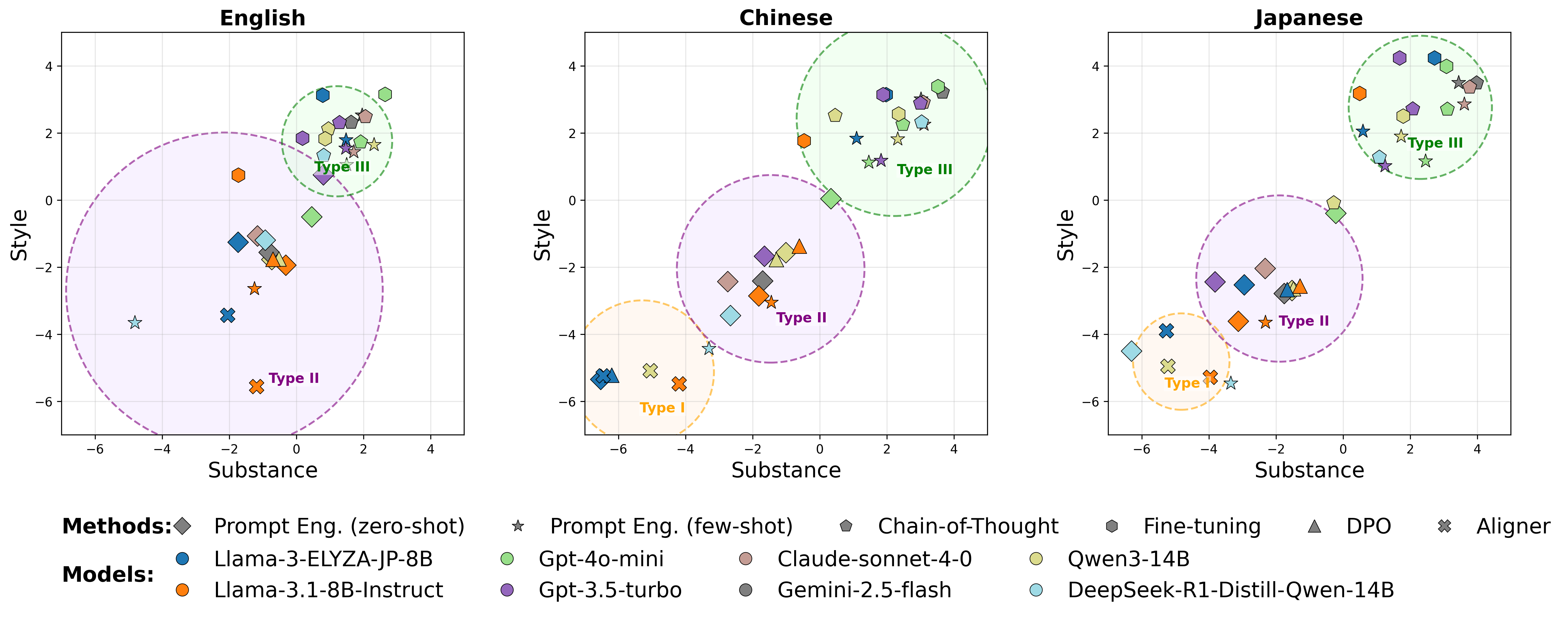}
    \caption{Normalized performance scores across English, Chinese, and Japanese users. The $X$-axis represents Substance metrics ($M1$: Word Mover's Distance, $M2$: Sentence Similarity, $M3$: BLEU); the $Y$-axis represents Style metrics ($M4$: Normalized Length Similarity, $M5$: Type-Token Ratio, $M6$: History Similarity). Colored circles denote automatically generated performance clusters.}
    \label{fig:score_plot}
\end{figure*}

Under each alignment method--LLM model, we evaluate personalized alignment separately for all 100 users and compute metric scores at the individual level. We then aggregate the results by averaging across users within each language group. The original (unnormalized) scores are reported in Table~5 in Appendix~D.

To enable fair comparison across different metrics, we apply z-score normalization (detail in Appendix D.) to each metric. The normalized results for M1--M6 are summarized in Table~2, where each cell reports the corresponding z-score. All scores are macro-averaged over participants within each language group (English: 33, Chinese: 34, Japanese: 33). The key findings based on these normalized results are discussed below.

\textbf{Advanced alignment methods} generally \textbf{outperform zero-shot baselines} across all linguistic contexts. As evidenced by the data in Table~\ref{tab:main-all-zscore}, while \textit{Prompt Eng. (zero-shot)} serves as a functional baseline, it consistently yields negative z-scores in critical metrics such as BLEU (M3) and Length Similarity (M4) across Chinese and Japanese. In contrast, methods such as \textit{Chain-of-Thought (CoT)} and \textit{Fine-tuning} shift the performance distribution significantly into the positive range, indicating that explicit reasoning steps and task-specific optimization are essential for achieving above-average persona consistency and linguistic quality.

\textbf{Few-shot prompting} provides a \textbf{substantial performance boost over zero-shot}, particularly in cross-lingual substance and style matching. The transition from zero-shot to \textit{Prompt Eng. (few-shot)} results in a marked increase in M3 (BLEU) scores; for instance, Gemini-2.5-flash improves from $-1.188$ to an orange-highlighted $1.557$ in English. This suggests that providing even a small number of context-rich examples allows models to better capture the ``Substance'' (M1--M3) required for accurate cross-lingual persona emulation, effectively anchoring the model's output to the desired target distribution.

\textbf{Chain-of-Thought (CoT)} serves as the \textbf{most robust method} for maintaining high-quality reasoning and history similarity in East Asian languages. In the Chinese and Japanese sections, CoT consistently secures top-three rankings across multiple metrics. Specifically, Gemini-2.5-flash and Claude-sonnet-4-0 achieve their highest Japanese M3 and M4 scores under this configuration. This finding implies that the intermediate reasoning steps inherent in CoT are vital for navigating the syntactic and cultural complexities of Chinese and Japanese, leading to superior ``Style'' (M4--M6) scores.

\textbf{Fine-tuning} emerges as the superior method for \textbf{optimizing structural metrics and linguistic diversity.} In the Japanese dataset, \textit{Fine-tuning} dominates the M4 (Length Similarity) and M5 (TTR) metrics, with Llama-3-ELYZA-JP-8B and Gpt-3.5-turbo achieving several ``Best'' and ``Second-best'' highlights. The high M5 scores (e.g., $1.587$ for Gpt-3.5-turbo) indicate that fine-tuned models produce more varied and sophisticated vocabulary compared to prompt-based methods, which often fall into repetitive patterns or overly standardized language.

\textbf{DPO and Aligner } methods show inconsistent results, often \textbf{struggling with stylistic consistency.} While DPO helps specific models like Llama-3.1-8B-Instruct achieve a second-best English M1 score ($0.385$), the \textit{Aligner} method shows significantly lower performance, often dipping into deep negative z-scores (e.g., $-3.494$ for Qwen3-14B in English M6). This suggests that post-hoc alignment without underlying reasoning or sufficient context may inadvertently diminish the persona's distinct characteristics, leading to outputs that are linguistically safe but stylistically bland.

\subsection{Clustering Analysis of Substance and Style}
To analyze the overall performance in terms of Substance and Style, we define \emph{Substance} as the sum of the z-scores of metrics $M1$, $M2$, and $M3$, and \emph{Style} as the sum of the z-scores of metrics $M4$, $M5$, and $M6$.
Based on these two dimensions, the classification of models into Type~I, Type~II, and Type~III clusters is performed using the K-Means clustering algorithm. As visualized in Figure~\ref{fig:score_plot}, the algorithm partitions the data into three regions based on Euclidean distance: Type~III (optimal performance, top-right), Type~II (baseline performance, center), and Type~I (lower-fidelity results, bottom-left), where key findings are shown as follows.

\textbf{The experimental results across all three languages demonstrate a clear stratified progression from low-fidelity clusters to high-performance persona emulation.} As illustrated in the scatter plots, the models and methods generally aggregate into three distinct zones: Type I (Low Substance, Low Style), Type II (Baseline Performance), and Type III (Optimal Alignment). This visualization confirms that the combination of Substance (metrics M1--M3) and Style (metrics M4--M6) is not uniform, but rather follows a predictable trajectory of improvement as alignment complexity increases.

\textbf{Type III clusters represent the state-of-the-art frontier, dominated by Chain-of-Thought and Fine-tuning across English, Chinese, and Japanese.} In all three subplots, the Type III zone (indicated by green dashed circles) contains the highest density of positive z-scores, particularly for the \textit{Chain-of-Thought} and \textit{Fine-tuning} methods. This clustering suggests that these techniques are uniquely capable of simultaneously optimizing for linguistic accuracy and stylistic persona consistency, pushing models toward the upper-right quadrant of the performance space.

\textbf{A distinct linguistic gap is visible in the distribution of Type I and Type II clusters between English and East Asian languages.} While the English results are more tightly concentrated within the Type II and Type III zones, the Chinese and Japanese plots show a more significant "tail" of models falling into the Type I category (orange dashed circles). This indicates that many baseline methods, particularly \textit{Aligner} and certain \textit{Prompt Eng. (zero-shot)} configurations, struggle significantly more with the stylistic nuances of East Asian personas, leading to a wider variance in cross-lingual performance.

\textbf{Model-specific clusters reveal that proprietary models like Gemini-2.5-flash and Gpt-4o-mini maintain superior stylistic stability across languages.} Despite changes in the underlying language, these specific models (represented by green and pink circles) consistently appear within or near the Type III clusters. This suggests that the internal alignment of these high-parameter models is robust enough to maintain persona "Style" even when the "Substance" is subject to the complexities of translation and cross-lingual cultural shifts.

\textbf{The transition from Type II to Type III is most frequently driven by the shift from zero-shot to few-shot and reasoning-based methodologies.} The cluster figures show that \textit{Prompt Eng. (few-shot)} data points (star markers) typically act as the bridge between baseline performance and high-fidelity output.Providing concrete examples shifts models from the diffuse Type~II region to the concentrated Type~III zone, highlighting that context and reasoning drive successful cross-lingual persona emulation.

\subsection{Leaderboard Analysis}

\begin{table}[t]
\centering
\caption{Leaderboard of top 20 configurations across English, Chinese, and Japanese.}
\label{tab:leaderboard_colored}
\small
\setlength{\tabcolsep}{2pt} % Tighten padding for single column
\begin{tabularx}{\columnwidth}{@{} p{0.4cm} >{\raggedright\arraybackslash}X l rrr @{}}
\toprule
\textbf{Rk} & \textbf{Alignment Method} & \textbf{Model} & \textbf{Sub} & \textbf{Sty} & \textbf{Ovr} \\ \midrule
1 & \cellcolor{colorLoRA} Fine-tuning & \texttt{gpt-4o-mini} & 1.03 & 1.17 & 1.10 \\
2 & \cellcolor{colorCoT} CoT & \texttt{gemini-2.5-flash} & 1.03 & 1.00 & 1.02 \\
3 & \cellcolor{colorCoT} CoT & \texttt{claude-sonnet-4-0} & 0.99 & 0.97 & 0.98 \\
4 & \cellcolor{colorFewShot} Few-shot & \texttt{gemini-2.5-flash} & 0.93 & 1.00 & 0.97 \\
5 & \cellcolor{colorLoRA} Fine-tuning & \texttt{L3-ELYZA} & 0.61 & 1.17 & 0.89 \\
6 & \cellcolor{colorFewShot} Few-shot & \texttt{claude-sonnet-4-0} & 0.94 & 0.73 & 0.83 \\
7 & \cellcolor{colorCoT} CoT & \texttt{gpt-3.5-turbo} & 0.71 & 0.88 & 0.79 \\
8 & \cellcolor{colorCoT} CoT & \texttt{gpt-4o-mini} & 0.83 & 0.74 & 0.79 \\
9 & \cellcolor{colorLoRA} Fine-tuning & \texttt{gpt-3.5-turbo} & 0.42 & 1.03 & 0.72 \\
10 & \cellcolor{colorLoRA} Fine-tuning & \texttt{Qwen3-14B} & 0.55 & 0.77 & 0.66 \\
\midrule
11 & \cellcolor{colorFewShot} Few-shot & \texttt{Qwen3-14B} & 0.71 & 0.60 & 0.65 \\
12 & \cellcolor{colorCoT} CoT & \texttt{DS-R1} & 0.55 & 0.55 & 0.55 \\
13 & \cellcolor{colorFewShot} Few-shot & \texttt{L3-ELYZA} & 0.35 & 0.63 & 0.49 \\
14 & \cellcolor{colorFewShot} Few-shot & \texttt{gpt-4o-mini} & 0.60 & 0.37 & 0.49 \\
15 & \cellcolor{colorFewShot} Few-shot & \texttt{gpt-3.5-turbo} & 0.51 & 0.42 & 0.46 \\
16 & \cellcolor{colorCoT} CoT & \texttt{Qwen3-14B} & 0.12 & 0.51 & 0.32 \\
17 & \cellcolor{colorLoRA} Fine-tuning & \texttt{Llama-3.1-8B} & -0.19 & 0.63 & 0.22 \\
18 & \cellcolor{white} Zero-shot & \texttt{gpt-4o-mini} & 0.06 & -0.09 & -0.02 \\
19 & \cellcolor{white} Zero-shot & \texttt{gpt-3.5-turbo} & -0.52 & -0.37 & -0.45 \\
20 & \cellcolor{colorDPO} DPO & \texttt{Llama-3.1-8B} & -0.29 & -0.63 & -0.46 \\
\bottomrule
\end{tabularx}
\end{table}

To compare the effectiveness of different alignment strategies across model architectures, we analyze a leaderboard of model configurations ranked by their Substance, Style, and overall performance and the key findings are shown as follows.

\textbf{Advanced alignment methods, specifically Fine-tuning and Chain-of-Thought (CoT), dominate the top of the leaderboard, consistently outperforming zero-shot baselines.} As shown in Table~\ref{tab:leaderboard_colored}, the top three spots are occupied by these methodologies, with \texttt{gpt-4o-mini} (Fine-tuning) and \texttt{gemini-2.5-flash} (CoT) achieving the highest overall z-scores of $1.10$ and $1.02$, respectively. This indicates that explicit stylistic optimization or structured reasoning steps are critical prerequisites for achieving high-fidelity persona emulation in a cross-lingual context.

\textbf{Top-tier proprietary models achieve a more balanced trade-off between Substance and Style than open-weight variants.} Models such as gemini-2.5-flash and claude-sonnet-4-0 show comparable performance across both dimensions (e.g., Gemini-2.5-flash with CoT: Substance $1.03$, Style $1.00$). In contrast, some open-weight models exhibit a pronounced style bias; for example, L3-ELYZA (Fine-tuning) ranks first in Style ($1.17$) but lags in Substance ($0.61$).

\textbf{The gpt-4o-mini and Gemini-2.5-flash models exhibit the highest versatility across different alignment techniques.} The \texttt{gpt-4o-mini} model appears in the top 20 using four distinct methods, Fine-tuning, CoT, Few-shot, and Zero-shot, while \texttt{gemini-2.5-flash} holds two of the top four positions. This highlights their robustness as base architectures for cross-lingual tasks, maintaining high performance regardless of whether the implementation utilizes prompting-based or parameter-efficient tuning strategies.

Further, we report some detailed \textbf{case studies} in Appendix~F.

\section{Conclusion}
We introduced YNTP-100, a benchmark for personalized response generation based on multi-day human--agent interactions. We also proposed the 2S Principle to evaluate personalized alignment. Our results show that current methods only partially capture user-specific patterns, and we hope YNTP-100 will support future work on scalable and privacy-aware personalization.

\section*{Limitations}

This work has several limitations. First, although YNTP-100 involves 100 participants and multi-day interactions, the scale remains modest compared to large-scale personalization scenarios, and future work could extend the benchmark to more users and longer time spans. Second, user responses are collected through a controlled human--agent interaction environment rather than real-world SNS or email platforms; while this design enables systematic data collection under privacy constraints, it may not fully capture all aspects of naturally occurring communication. Third, while the proposed evaluation metrics capture complementary aspects of content and style, automatic metrics may not fully reflect human judgments of personalization quality, and incorporating human evaluation remains an important direction for future work.

% T

%%%%%%%%%%%%%%%%%%%%%%%%%%%%%%%%%%%%%%%%%%%%%%%%%%%%%%%%%%%%%%%%%%%%%%%%

%%% The acknowledgments section is defined using the "acks" environment
%%% (rather than an unnumbered section). The use of this environment 
%%% ensures the proper identification of the section in the article 
%%% metadata as well as the consistent spelling of the heading.

% \begin{acks}
% If you wish to include any acknowledgments in your paper (e.g., to 
% people or funding agencies), please do so using the `\texttt{acks}' 
% environment. Note that the text of your acknowledgments will be omitted
% if you compile your document with the `\texttt{anonymous}' option.
% \end{acks}

%%%%%%%%%%%%%%%%%%%%%%%%%%%%%%%%%%%%%%%%%%%%%%%%%%%%%%%%%%%%%%%%%%%%%%%%

%%% The next two lines define, first, the bibliography style to be 
%%% applied, and, second, the bibliography file to be used.

% \bibliographystyle{acl_natbib} 
\bibliography{sample}

%%%%%%%%%%%%%%%%%%%%%%%%%%%%%%%%%%%%%%%%%%%%%%%%%%%%%%%%%%%%%%%%%%%%%%%%

\appendix
\section*{Appendix}
\section{Existing Dataset for Personlized Alignment}
\begin{table*}[t]
\centering
\caption{Summary of key personalization-related datasets.}
\label{tab:personalization-datasets}
\small
\setlength{\tabcolsep}{3.5pt}
\begin{tabular}{p{1.5cm} p{2.5cm} p{2.5cm} p{2.5cm} p{2.5cm} p{2cm}}
\toprule
\textbf{Dataset} & \textbf{Input} & \textbf{Output} & \textbf{History/Persona} & \textbf{Size (Train/Dev)} & \textbf{Metric} \\
\midrule
LaMP (1--7) & Multi-domain prompts (papers, movies, reviews, news, tweets) & Personalized or preference-aware outputs (titles, tags, ratings, headlines, paraphrases) & User-specific histories (authored papers, rated items, past headlines, tweets, etc.) & 6k--20k per task & Accuracy, F1, MAE, ROUGE-1/L \\
LongLaMP & Long-form texts (reviews, blogs) & Personalized long-form outputs & User’s prior writings & $\sim$10k examples & BLEU, ROUGE, METEOR \\
P-SOUPS & Prompts across news/dialogue/review domains & Personalized responses & User profiles + multi-task history & $\sim$1M examples (2024) & BLEU, ROUGE, human eval. \\
PRISM & Conversational context (chat logs) & Next-turn personalized response & User style embeddings + history & $\sim$50k conversations & BLEU, StyleSim, human eval. \\
CUSTOM & Email prompts & Personalized email response & Few-shot (4 train, 1 test per user) & 2 prompts × 2 demos + 1 test per user & Human preference, style match \\
PersonalLLM & Generic task prompt & Personalized response & Learned user embeddings & $\sim$100k samples & BLEU, style match, human eval. \\
PERSONA & Task prompt + persona card & Persona-consistent response & Explicit persona description & $\sim$60k examples & StyleSim, human eval. \\
\midrule
FLASK & Task + user preference & Aligned model response & Human-annotated preferences & $\sim$90k examples & BLEU, ROUGE, human eval. \\
REGEN & Dialogue context & Personalized continuation & Past dialogues with evolving preferences & $\sim$50k dialogues & Consistency, human eval. \\
ALOE & Prompt + rationale & Personalized response + explanation & User-specific rationales & $\sim$80k examples & BLEU, ROUGE, faithfulness \\
PREFEVAL & Candidate generations & Human preference ranking & Annotator preference judgments & $\sim$20k comparisons & Win-rate, agreement \\
\bottomrule
\end{tabular}
\end{table*}

Table~\ref{tab:personalization-datasets} summarizes representative datasets used in prior work on personalized alignment and preference modeling. These datasets vary widely in their input–output formats, sources of personalization signals, scale, and evaluation protocols. Early benchmarks such as LaMP and LongLaMP focus on task-oriented personalization conditioned on user histories, while later datasets (e.g., P-SOUPS, PRISM, and PersonalLLM) explore multi-domain personalization and conversational settings using user profiles or learned embeddings. Other datasets, including FLASK, REGEN, ALOE, and PREFEVAL, emphasize preference-based alignment through human feedback rather than explicit user identity modeling. Together, these datasets provide the foundation for studying personalization and alignment, but they primarily target static or task-specific customization rather than modeling how individuals respond in natural, multi-turn communication over time.

\section{Data Collection Protocol}

\subsection{Instructions Given to Participants}
Participants were provided with written instructions prior to taking part in the study. They were asked to access a web-based experimental platform and complete a multi-day human--agent dialogue task according to the guidelines. The instructions emphasized that the interaction was part of a research study and that participants should avoid sharing any personally identifiable, private, or sensitive information during the dialogues. Participants were also informed of the submission procedure upon completion of the task and were instructed to follow the specified steps for finalizing their participation.

\subsection{Recruitment and Payment}
Participants were recruited on a voluntary basis through direct invitation. No obligation to participate was imposed, and individuals who chose not to take part were not required to provide a reason. Participants who completed the experiment and submitted their results as instructed received a small monetary incentive in the form of an electronic gift card. Compensation was provided after confirmation of successful submission and was not contingent on the content or quality of the responses.

\subsection{Data Consent}
Before participating, individuals were explicitly informed that the dialogue data collected during the experiment could be used for research purposes, including the analysis, training, and evaluation of AI models. They were also notified that portions of the data might be released publicly as part of a research dataset, in anonymized form. Participation was taken as informed consent to these terms. No personally identifiable information was collected, and all released data were processed to remove potential identifiers prior to analysis or publication.

\section{Metric Equations}

Let $y$ and $\hat{y}$ denote the reference (ground-truth) and predicted responses for a user $u$, respectively, and let $H_u$ denote the set of historical responses from user $u$. We use fixed pre-trained encoders to obtain word and sentence embeddings, and $\cos(\cdot,\cdot)$ denotes cosine similarity.

\textbf{M1 (Word Mover’s Distance).}
This metric measures semantic dissimilarity between predicted and reference responses by computing the minimum transport cost between their word embeddings. Lower values indicate closer semantic alignment:
\begin{align*}
     \mathrm{WMD} &= \min_{T \geq 0} \sum_{i=1}^n \sum_{j=1}^n T_{ij} \lVert x_i - x_j \rVert_2 \\
     \text{s.t. } \sum_{j=1}^n T_{ij} &= d_i,\quad \forall i \in \{1,\dots,n\}, \\
     \sum_{i=1}^n T_{ij} &= d'_j,\quad \forall j \in \{1,\dots,n\}.
\end{align*}
Here, $x_i$ and $x_j$ denote the word embeddings of the $i$-th word in the predicted response and the $j$-th word in the reference response, respectively; $n$ is the number of unique words in the union of both responses; $T_{ij}$ is the transport flow from word $i$ to word $j$; and $d_i$ and $d'_j$ are the normalized word frequency distributions of the predicted and reference responses. $\lVert\cdot\rVert_2$ denotes the Euclidean norm.

\textbf{M2 (Sentence Similarity).}
This metric evaluates sentence-level semantic closeness using cosine similarity between sentence embeddings:
\[
\mathrm{SentSim}(y, \hat{y}) = \cos(\mathbf{e}_y, \mathbf{e}_{\hat{y}}),
\]
where $\mathbf{e}_y$ and $\mathbf{e}_{\hat{y}}$ are sentence embeddings of the reference and predicted responses, respectively, obtained from a fixed pre-trained encoder.

\textbf{M3 (BLEU).}
This metric quantifies surface-level content overlap between predicted and reference responses using n-gram precision with a brevity penalty:
\begin{align*}
\mathrm{BLEU} &= BP \times \exp\!\Bigg(\sum_{n=1}^{N} w_n \log P_n\Bigg), \\
P_n &= \frac{\mathrm{Count}_{\mathrm{clip}}(\text{ngram}_{\mathrm{Pred}})}{\mathrm{Count}(\text{ngram}_{\mathrm{Ref}})}, \\
BP &=
\begin{cases}
    1, & \text{if } len_{\mathrm{Pred}} > len_{\mathrm{Ref}}, \\
    e^{1 - len_{\mathrm{Ref}} / len_{\mathrm{Pred}}}, & \text{otherwise}.
\end{cases}
\end{align*}
Here, $N$ is the maximum n-gram order, $w_n$ are uniform weights, $P_n$ is the modified n-gram precision, $\mathrm{Count}_{\mathrm{clip}}(\cdot)$ denotes clipped n-gram counts, $\text{ngram}_{\mathrm{Pred}}$ and $\text{ngram}_{\mathrm{Ref}}$ are the n-grams from the predicted and reference responses, and $len_{\mathrm{Pred}}$ and $len_{\mathrm{Ref}}$ denote their token lengths.

\emph{Style metrics (M4--M6)} assess stylistic consistency and user-specific expression patterns.

\textbf{M4 (Normalized Length Similarity).}
This metric measures verbosity alignment by computing the ratio between the shorter and longer response lengths:
\begin{align*}
     \mathrm{NLS}
     = \frac{\min\{len_{\mathrm{Pred}},\, len_{\mathrm{Ref}}\}}{\max\{len_{\mathrm{Pred}},\, len_{\mathrm{Ref}}\}}.
\end{align*}
Here, $len_{\mathrm{Pred}}$ and $len_{\mathrm{Ref}}$ denote the number of tokens in the predicted and reference responses, respectively.

\textbf{M5 (Type--Token Ratio).}
This metric evaluates lexical richness as the proportion of unique tokens relative to the total number of tokens:
\begin{align*}
      \mathrm{TTR} = \frac{V}{N},
\end{align*}
where $V$ denotes the number of distinct tokens and $N$ the total token count in the response.

\textbf{M6 (History Similarity).}
This metric measures long-term stylistic consistency by computing the average embedding similarity between the generated response and a user’s historical responses:
\[
\mathrm{HistSim}(\hat{y}, H_u)
= \frac{1}{|H_u|} \sum_{h \in H_u} \cos(\mathbf{e}_{\hat{y}}, \mathbf{e}_h),
\]
where $H_u$ denotes the set of historical responses from user $u$, each $h \in H_u$ is a past response, and $\mathbf{e}_{\hat{y}}$ and $\mathbf{e}_h$ are their corresponding sentence embeddings.

\section{Original Experiment Results}
The original results corresponding to Table~2 are listed in Table~5.
\subsection{Z-score Normalization}

Because the six evaluation metrics (M1--M6) differ in scale and direction, we apply z-score normalization to enable fair comparison across methods and models. Let $s^{(m)}_{u}$ denote the raw score of metric $m \in \{1,\dots,6\}$ for user $u$, where $s^{(m)}_{u}$ is computed from the generated response $\hat{y}$ and the reference response $y$. For each metric $m$, we first compute the mean and standard deviation over all users in the evaluation set:
\begin{align*}
\mu^{(m)} &= \frac{1}{U} \sum_{u=1}^{U} s^{(m)}_{u}, \\
\sigma^{(m)} &= \sqrt{\frac{1}{U} \sum_{u=1}^{U} \left(s^{(m)}_{u} - \mu^{(m)}\right)^2},
\end{align*}
where $U$ is the total number of users.

We then normalize each individual score using:
\[
z^{(m)}_{u} = \frac{s^{(m)}_{u} - \mu^{(m)}}{\sigma^{(m)}}.
\]
This transformation rescales each metric to have zero mean and unit variance, making scores comparable across metrics with different ranges and distributions.

For each method--model combination, we report macro-averaged z-scores by averaging $z^{(m)}_{u}$ over users within the same language group. As a result, each cell in Table~2 represents the mean normalized performance of a method on a given metric for that language. The original (unnormalized) scores are provided in Table~5 in Appendix~D for completeness.

\begin{table*}[htbp]
\centering
\caption{
The original results (before z-score normalization) across three languages corresponding to Table 2: macro-averaged scores over 33 English users, 34 Chinese users, and 33 Japanese users. Each cell represents the average score of all participants in that language group on the
corresponding metric. Lower values of M1 indicate better distance alignment, while higher values of M2–M6
indicate stronger similarity or consistency.
}
\label{tab:main-all}
\small
\setlength{\tabcolsep}{1.5pt}
\renewcommand{\arraystretch}{0.95}

% ===================== key line added =====================
\resizebox{\textwidth}{!}{
% ===========================================================

\begin{tabular}{
p{1.5cm} p{1.5cm} 
r r r r r r | 
r r r r r r | 
r r r r r r
}
\toprule
\multirow{2}{*}{\textbf{Method}} & \multirow{2}{*}{\textbf{Base Model}} &
\multicolumn{6}{c|}{\textbf{English (33 users)}} &
\multicolumn{6}{c|}{\textbf{Chinese (34 users)}} &
\multicolumn{6}{c}{\textbf{Japanese (33 users)}} \\
\cmidrule(lr){3-8} \cmidrule(lr){9-14} \cmidrule(lr){15-20}
 &  & M1$\downarrow$ & M2$\uparrow$ & M3$\uparrow$ & M4$\uparrow$ & M5$\uparrow$ & M6$\uparrow$
    & M1$\downarrow$ & M2$\uparrow$ & M3$\uparrow$ & M4$\uparrow$ & M5$\uparrow$ & M6$\uparrow$
    & M1$\downarrow$ & M2$\uparrow$ & M3$\uparrow$ & M4$\uparrow$ & M5$\uparrow$ & M6$\uparrow$ \\
\midrule
\multirow{6}{=}{\parbox{2.0cm}{\raggedright Prompt Eng. (zero-shot)}}
  & Gpt-3.5-turbo & 0.242 & 0.427 & 0.0104 & 0.337 & 0.799 & 0.430 
                  & 1.042 & 0.417 & 0.00839 & 0.252 & 0.672 & 0.461 
                  & 0.999 & 0.178 & 0.0076 & 0.252 & 0.541 & 0.289 \\
  & Gpt-4o-mini   & 0.242 & 0.425 & 0.00833 & 0.236 & 0.750 & 0.412 
                  & 0.928 & 0.538 & 0.00765 & 0.226 & 0.692 & 0.723 
                  & 0.717 & 0.277 & 0.0138 & 0.236 & 0.564 & 0.582 \\
  & Gemini-2.5-flash
                  & 0.244 & 0.389 & 0.00353 & 0.121 & 0.605 & 0.426 
                  & 0.982 & 0.406 & 0.00288 & 0.072 & 0.496 & 0.631 
                  & 0.751 & 0.188 & 0.00576 & 0.086 & 0.414 & 0.483 \\
  & Claude-sonnet-4-0
                  & 0.235 & 0.367 & 0.00281 & 0.105 & 0.665 & 0.436 
                  & 1.042 & 0.351 & 0.00281 & 0.098 & 0.568 & 0.552 
                  & 0.812 & 0.177 & 0.00610 & 0.131 & 0.511 & 0.470 \\
  & DeepSeek-R1-Distill-Qwen-14B & 0.232 & 0.371 & 0.0038 & 0.075 & 0.543 & 0.465
               & 1.057 & 0.373 & 0.0037 & 0.047 & 0.473 & 0.515
               & 1.205 & 0.140 & 0.0023 & 0.037 & 0.381 & 0.298 \\
  & Qwen3-14B & 0.241 & 0.390 & 0.0038 & 0.128 & 0.590 & 0.420
              & 0.942 & 0.413 & 0.0045 & 0.110 & 0.567 & 0.671
              & 0.734 & 0.189 & 0.0066 & 0.112 & 0.409 & 0.480 \\
  & Llama-3.1-8B-Instruct & 0.238 & 0.395 & 0.0059 & 0.145 & 0.535 & 0.422
                          & 1.001 & 0.386 & 0.0049 & 0.126 & 0.422 & 0.578
                          & 0.887 & 0.144 & 0.0067 & 0.145 & 0.345 & 0.375 \\
  & Llama-3-ELYZA-JP-8B & 0.308 & 0.328 & 0.0068 & 0.312 & 0.778 & 0.358
                        & 1.230 & 0.117 & 0.0014 & 0.116 & 0.423 & 0.213
                        & 0.902 & 0.160 & 0.0091 & 0.216 & 0.467 & 0.370 \\
\midrule
\multirow{6}{=}{\parbox{2.0cm}{\raggedright Prompt Eng. (few-shot)}}
  & Gpt-3.5-turbo   & 0.249 & 0.437 & 0.0142 & 0.382 & 0.830 & 0.445
                    & 0.920 & 0.606 & 0.0161 & 0.340 & 0.755 & 0.753
                    & 0.717 & 0.450 & 0.0189 & 0.367 & 0.637 & 0.624 \\
  & Gpt-4o-mini   & 0.247 & 0.451 & 0.0130 & 0.368 & 0.831 & 0.428 
                  & 0.907 & 0.615 & 0.0111 & 0.327 & 0.769 & 0.745
                  & 0.688 & 0.464 & 0.0287 & 0.349 & 0.639 & 0.657 \\
  & Gemini-2.5-flash
                  & 0.276 & 0.411 & 0.0213 & 0.621 & 0.901 & 0.417 
                  & 0.912 & 0.628 & 0.0242 & 0.536 & 0.859 & 0.789 
                  & 0.694 & 0.558 & 0.0342 & 0.576 & 0.748 & 0.725 \\
  & Claude-sonnet-4-0
                  & 0.246 & 0.433 & 0.0160 & 0.426 & 0.824 & 0.433 
                  & 0.888 & 0.629 & 0.0226 & 0.436 & 0.821 & 0.786 
                  & 0.683 & 0.540 & 0.0362 & 0.507 & 0.717 & 0.714 \\
  & DeepSeek-R1-Distill-Qwen-14B & 0.236 & 0.124 & 0.0017 & 0.016 & 0.360 & 0.420
               & 1.062 & 0.325 & 0.0013 & 0.012 & 0.333 & 0.502
               & 0.825 & 0.069 & 0.0030 & 0.012 & 0.249 & 0.292 \\
  & Qwen3-14B & 0.246 & 0.439 & 0.0193 & 0.478 & 0.830 & 0.429
              & 0.914 & 0.618 & 0.0190 & 0.416 & 0.795 & 0.756
              & 0.706 & 0.470 & 0.0218 & 0.452 & 0.666 & 0.661 \\
  & Llama-3.1-8B-Instruct & 0.269 & 0.356 & 0.0051 & 0.145 & 0.260 & 0.457
                          & 0.962 & 0.379 & 0.0046 & 0.145 & 0.297 & 0.630
                          & 0.806 & 0.174 & 0.0057 & 0.168 & 0.270 & 0.419 \\
  & Llama-3-ELYZA-JP-8B & 0.284 & 0.385 & 0.0211 & 0.541 & 0.878 & 0.410
                        & 0.975 & 0.608 & 0.0149 & 0.449 & 0.788 & 0.738
                        & 0.796 & 0.471 & 0.0178 & 0.491 & 0.692 & 0.628 \\
\midrule
\multirow{6}{=}{\parbox{2.0cm}{\raggedright  Chain-of-Thought}}
  & Gpt-3.5-turbo & 0.250 & 0.428 & 0.0139 & 0.519 & 0.884 & 0.434
                  & 0.909 & 0.652 & 0.0224 & 0.519 & 0.849 & 0.792
                  & 0.722 & 0.534 & 0.0231 & 0.497 & 0.718 & 0.699 \\
  & Gpt-4o-mini   & 0.246 & 0.442 & 0.0164 & 0.467 & 0.849 & 0.430
                  & 0.900 & 0.651 & 0.0171 & 0.439 & 0.821 & 0.782
                  & 0.691 & 0.542 & 0.0312 & 0.483 & 0.706 & 0.720 \\
  & Gemini-2.5-flash & 0.274 & 0.418 & 0.0184 & 0.591 & 0.893 & 0.417
                     & 0.907 & 0.643 & 0.0283 & 0.555 & 0.876 & 0.790
                     & 0.687 & 0.551 & 0.0401 & 0.571 & 0.746 & 0.731 \\
  & Claude-sonnet-4-0  & 0.253 & 0.438 & 0.0182 & 0.566 & 0.868 & 0.435
                       & 0.893 & 0.633 & 0.0228 & 0.523 & 0.850 & 0.791
                       & 0.685 & 0.557 & 0.0371 & 0.542 & 0.743 & 0.738 \\
  & DeepSeek-R1-Distill-Qwen-14B & 0.258 & 0.416 & 0.0124 & 0.442 & 0.835 & 0.422
                & 0.892 & 0.634 & 0.0221 & 0.463 & 0.813 & 0.780
                & 0.731 & 0.424 & 0.0200 & 0.397 & 0.651 & 0.626 \\
  & Qwen3-14B & 0.290 & 0.390 & 0.0176 & 0.584 & 0.907 & 0.407
              & 0.976 & 0.598 & 0.0100 & 0.499 & 0.879 & 0.730
              & 0.797 & 0.391 & 0.0134 & 0.349 & 0.514 & 0.585 \\
\midrule
\multirow{1}{=}{\parbox{2.0cm}{\raggedright Fine-tuning}}
  & Gpt-3.5-turbo  & 0.302 & 0.353 & 0.0167 & 0.586 & 0.899 & 0.398
                   & 0.890 & 0.630 & 0.0174 & 0.561 & 0.878 & 0.774
                   & 0.746 & 0.562 & 0.0198 & 0.597 & 0.837 & 0.738 \\
  & Gpt-4o-mini    & 0.274 & 0.380 & 0.0284 & 0.630 & 0.864 & 0.449
                   & 0.874 & 0.648 & 0.0281 & 0.595 & 0.870 & 0.789
                   & 0.729 & 0.567 & 0.0334 & 0.599 & 0.785 & 0.746 \\
  & Qwen3-14B & 0.295 & 0.366 & 0.0195 & 0.519 & 0.871 & 0.418
              & 0.937 & 0.627 & 0.0210 & 0.503 & 0.838 & 0.765
              & 0.730 & 0.503 & 0.0227 & 0.472 & 0.707 & 0.697 \\
  & Llama-3.1-8B-Instruct  & 0.296 & 0.302 & 0.0086 & 0.503 & 0.880 & 0.375
                           & 1.011 & 0.575 & 0.0068 & 0.410 & 0.838 & 0.720
                           & 0.789 & 0.522 & 0.0127 & 0.505 & 0.780 & 0.706 \\
  & Llama-3-ELYZA-JP-8B & 0.273 & 0.349 & 0.0193 & 0.635 & 0.853 & 0.450 
                & 0.941 & 0.641 & 0.0173 & 0.561 & 0.855 & 0.791
                & 0.737 & 0.571 & 0.0294 & 0.621 & 0.793 & 0.760 \\
\midrule
\multirow{1}{=}{\parbox{2.0cm}{\raggedright DPO}}
  & Qwen3-14B & 0.242 & 0.389 & 0.0054 & 0.122 & 0.585 & 0.423
              & 0.949 & 0.392 & 0.0038 & 0.101 & 0.561 & 0.654
              & 0.719 & 0.176 & 0.0065 & 0.113 & 0.421 & 0.476\\
  & Llama-3.1-8B-Instruct & 0.233 & 0.382 & 0.0043 & 0.113 & 0.559 & 0.430
                          & 0.935 & 0.448 & 0.0053 & 0.158 & 0.542 & 0.682
                          & 0.723 & 0.196 & 0.0078 & 0.160 & 0.381 & 0.487 \\
  & Llama-3-ELYZA-JP-8B  & 1.015 & 0.106 & 0.0070 & 0.297 & 0.914 & 0.273
                 & 1.179 & 0.093 & 0.0007 & 0.104 & 0.382 & 0.274
                 & 0.771 & 0.192 & 0.0082 & 0.162 & 0.434 & 0.423 \\

\midrule
\multirow{1}{=}{\parbox{2.0cm}{\raggedright Aligner}}
  & Qwen3-14B & 0.245 & 0.274 & 0.0020 & 0.037 & 0.295 & 0.269
                & 1.190 & 0.280 & 0.0011 & 0.028 & 0.294 & 0.422
                & 1.067 & 0.117 & 0.0025 & 0.053 & 0.286 & 0.303 \\
  & Llama-3.1-8B-Instruct  & 0.251 & 0.389 & 0.0016 & 0.037 & 0.261 & 0.361
                           & 1.137 & 0.320 & 0.0012 & 0.022 & 0.168 & 0.466
                           & 0.930 & 0.128 & 0.0026 & 0.042 & 0.185 & 0.354 \\
  & Llama-3-ELYZA-JP-8B  & 0.374 & 0.339 & 0.0076 & 0.234 & 0.706 & 0.303
                 & 1.220 & 0.125 & 0.0006 & 0.086 & 0.402 & 0.269
                 & 1.078 & 0.092 & 0.0047 & 0.106 & 0.426 & 0.292 \\

\bottomrule
\end{tabular}

} % end resizebox

\end{table*}

\section{Prompts and Hyperparameter for Each Method}

This section provides the exact prompts and hyperparameters used for each method.

% ===========================
% Zero-shot
% ===========================
\paragraph{Prompt Engineering (Zero-shot)}
\textbf{Training / inference method.}
In this setting, the model predicts the user’s response without any additional personalization signals. The model receives only the input question, without access to user history or auxiliary information.

\textbf{Prompt format.}
The user prompt contains the following placeholder:

\begin{tcolorbox}[colback=white,colframe=black!50,title=Zero-shot Prompt]
{\small
\begin{verbatim}
system prompt:
You have joined the share house as a new resident.
DaVinci, Donatello, Michelangelo, and Raffaello
are members of the share house. Please predict
your next response.

user prompt:
Please predict your response to the following
message:{question}
\end{verbatim}
}
\end{tcolorbox}

\begin{tcolorbox}[colback=gray!3,colframe=black!40,title=Zero-shot Hyperparameters (JSON)]
{\small
\begin{verbatim}
{
  "max_tokens": 512,
  "temperature": 0.7,
  "top_p": 1.0,
  "frequency_penalty": 0.0,
  "presence_penalty": 0.0
}
\end{verbatim}
}
\end{tcolorbox}

% ===========================
% Few-shot
% ===========================
\paragraph{Prompt Engineering (Few-shot)}
\textbf{Training / inference method.}
In this approach, the model predicts the user’s response based on examples from Day~1 to Day~4 interactions. These historical responses are included in the prompt, and the system instructions explicitly encourage the model to imitate the user’s communication style.

\textbf{Prompt format.}
The user prompt contains the following placeholders:

\begin{tcolorbox}[colback=white,colframe=black!50,title=Few-shot Prompt]
{\small
\begin{verbatim}
system prompt:
You have joined the share house as a new resident.
DaVinci, Donatello, Michelangelo, and Raffaello
are members of the share house. I will provide
you with the previous exchanges from the
conversation. Here, "A" refers to your reply.
Please carefully observe the tone, attitude,
values, and other cues. For example, pay
attention to the following points:
    - which first-person pronoun you use,
    - whether you tend to be concise or prefer
      detailed and polite expressions,
    - whether you use casual or formal language,
    - how long you usually make your responses,
    - how you use punctuation.
    Especially, please pay attention to the length
    of your responses.
    Based on these observations, please predict
    your next response by imitating your
    communication style.
    
user prompt:
previous interactions :
    {train_data}
    Please predict your response to the following
    message:{question}
\end{verbatim}
}
\end{tcolorbox}

\begin{tcolorbox}[colback=gray!3,colframe=black!40,title=Few-shot Hyperparameters (JSON)]
{\small
\begin{verbatim}
{
  "max_tokens": 512,
  "temperature": 0.7,
  "top_p": 1.0,
  "frequency_penalty": 0.0,
  "presence_penalty": 0.0
}
\end{verbatim}
}
\end{tcolorbox}

% ===========================
% Few-shot + MBTI inference
% ===========================
\paragraph{Prompt Engineering (Few-shot + MBTI Inference)}
\textbf{Training / inference method.}
This method conditions the model not only on few-shot examples from Day~1 to Day~4, but also on inferred MBTI personality information derived from the same interaction history.

\textbf{Input components.}
The user prompt contains the following placeholders:
\begin{itemize}
    \item \texttt{mbti\_info}:  
    The inferred MBTI profile of the user, represented as continuous ratios for each dimension (E/I, S/N, T/F, J/P).
    \item \texttt{train\_data}:  
    Conversation history from Day~1 to Day~4, used as few-shot examples.
    \item \texttt{question}:  
    The input message from Day~5 for which the model predicts the user’s response.
\end{itemize}

\begin{tcolorbox}[colback=white,colframe=black!50,title=Few-shot + MBTI Inference Prompt]
{\small
\begin{Verbatim}[breaklines, breakanywhere]
system prompt:
You have joined the share house as a new resident. DaVinci, Donatello,
Michelangelo, and Raffaello are members of the share house. I will provide
you with the previous exchanges from the conversation and your MBTI
information. Here, "A" refers to your reply. Please carefully observe the
tone, attitude, values, and other cues.

Based on these observations and the given MBTI information, please predict
your next response by imitating your communication style.

user prompt:
your MBTI information :
{mbti_info}
previous interactions :
{train_data}
Please predict your response to the following message:{question}
\end{Verbatim}
}
\end{tcolorbox}

\begin{tcolorbox}[colback=gray!3,colframe=black!40,title=MBTI Inference Hyperparameters (JSON)]
{\small
\begin{verbatim}
{
  "max_tokens": 512,
  "temperature": 0.6,
  "top_p": 1.0,
  "frequency_penalty": 0.0,
  "presence_penalty": 0.0
}
\end{verbatim}
}
\end{tcolorbox}

% ===========================
% CoT
% ===========================
\paragraph{Prompt Engineering (Few-shot + Latent CoT)}
\textbf{Training / inference method.}
This approach predicts the user’s response by conditioning on few-shot examples and latent communication traits inferred from prior interactions. The process is decomposed into two stages: (i) style inference and (ii) final response generation.

\textbf{Latent style inference.}
The model first infers latent stylistic attributes (tone, length, emotion, and interpersonal distance) from the user’s previous responses only. This step is used solely to guide generation and is not exposed in the output.

\textbf{Final response generation.}
The inferred style constraints are injected into the system prompt to guide the final response. The model is instructed to perform internal reasoning while suppressing intermediate explanations.

[Prompts and hyperparameters unchanged from your original version for reproducibility.]

% ===========================
% PEFT (LoRA)
% ===========================
\paragraph{PEFT (LoRA)}
\textbf{Training / inference method.}
Low-Rank Adaptation (LoRA) fine-tunes the model parameters using user-specific data.

[Prompts and hyperparameters unchanged.]

% ===========================
% DPO
% ===========================
\paragraph{Direct Preference Optimization (DPO)}
\textbf{Training method.}
DPO fine-tunes LLMs using preference pairs without requiring an explicit reward model. Using data from Day~1 to Day~4, we construct training triples as:
\begin{itemize}
  \item \textbf{Prompt}: the NPC question.
  \item \textbf{Rejected}: the model’s pre-training output.
  \item \textbf{Chosen}: the ground-truth user response.
\end{itemize}

[Prompts and hyperparameters unchanged.]

% ===========================
% Aligner
% ===========================
\paragraph{Aligner}
\textbf{Training method.}
Instead of directly fine-tuning the LLM, Aligner trains a sequence-to-sequence model to revise the LLM output. The base LLM parameters are frozen. Training uses triplets:
\begin{itemize}
  \item \textbf{Question}: NPC prompt.
  \item \textbf{Answer}: original model output.
  \item \textbf{Correction}: revised output generated by GPT-5-mini conditioned on Day~1--Day~4 data.
\end{itemize}

\paragraph{Augmented Aligner}
\textbf{Training method.}
To address data sparsity, the training set is augmented with 600 additional questions generated by GPT-5-mini.

\section{Case Study: Representative Examples of Prediction Performance}

\subsection{Case 1: Complete Style Mismatch - Extreme Brevity vs. Verbosity}

\textbf{Correct Response}: 
\begin{quote}
\texttt{"Yes. I agree."}
\end{quote}

\textbf{Predicted Response}: 
\begin{quote}
\texttt{"Let's make it easy!"}
\end{quote}

The metrics reveal a fundamental mismatch: WMD (M1) of 0.652 indicates high semantic distance despite both expressing agreement, while sentence similarity (M2) of 0.081 shows very low contextual similarity. Content similarity (M3) scores only 1/5 for minimal factual alignment, length similarity (M4) shows moderate mismatch at 0.684 (3 vs 4 words), and style similarity (M5) achieves only 2/5. This example perfectly illustrates how the model fails to capture extreme brevity, adding unnecessary enthusiasm (``Let's make it easy!'') where the user simply states ``Yes. I agree.''

\subsection{Case 2: Content Misalignment - Missing Personal Context}

\textbf{Correct Response}: 
\begin{quote}
\texttt{"let's see... I think i alway appreciate when i meet someone even my family. cause every moment never come back...especially my friends, family, boyfriend i cherish have own lifetime, but they give their time to spend with me. so i always feel happy when i meet important n precious people."}
\end{quote}

\textbf{Predicted Response}: 
\begin{quote}
\texttt{"oh, i see what you mean. so, like if we're at the party, we should take many photos and make unforgettable n unforgettable moments! and after the party, we should look at these photos and remember n remember our cool n fun time! is my response correct?"}
\end{quote}

Despite achieving perfect style similarity (M5: 5/5) and good length match (M4: 0.872), the response completely misses the user's deeply personal reflection. The moderate WMD (M1: 0.209) and partial content overlap (M3: 3/5) mask the fundamental misunderstanding—the model provides superficial suggestions about taking photos at parties instead of acknowledging the user's philosophical perspective on cherishing time with loved ones. History similarity (M6: 0.461/0.268) remains poor, indicating inconsistent personalization.

\subsection{Case 3: Successful Prediction - Analytical Alignment}

\textbf{Correct Response}: 
\begin{quote}
\texttt{"Well...if we have limited space in our flat, I guess I'd prefer furniture that is more multifunctional and space saving. That's life. However though, I think we could try to blast our creativity in other parts like decorations on the walls where we don't need to think about practicality."}
\end{quote}

\textbf{Predicted Response}: 
\begin{quote}
\texttt{"Well, I think I would prefer furniture that is more multi-functional and space-saving - I believe furniture that serves more than one purpose is more efficient and reduces waste. Although I do think we can still consider the same design principle even if we wanted to focus on design and aesthetic only."}
\end{quote}

This represents the model's best performance with WMD (M1) of 0.128 showing low semantic distance, sentence similarity (M2) of 0.714 indicating high contextual alignment, and content similarity (M3) of 4/5 demonstrating strong factual match. Length similarity (M4) achieves excellent alignment at 0.954, while style similarity (M5) scores 4/5. Both responses start with ``Well...'', express preference for multi-functional furniture, acknowledge the trade-off between practicality and aesthetics, and use similar analytical reasoning structures. The success occurs because the user's natural style aligns with the model's tendency toward analytical responses.

\subsection{Case 4: Context Misunderstanding - Wrong Conversation Thread}

\textbf{Correct Response}: 
\begin{quote}
\texttt{"yep making routine respectively is good for ours/"}
\end{quote}

\textbf{Predicted Response}: 
\begin{quote}
\texttt{"yeah true."}
\end{quote}

This example shows how misleading individual metrics can be. Despite achieving perfect style similarity (M5: 5/5) for matching casual tone, the predicted response completely misses the user's point about ``making routine respectively.'' High WMD (M1: 0.588), very low sentence similarity (M2: 0.139), severe length mismatch (M4: 0.204), and poor content match (M3: 2/5) reveal the model captures surface-level style while losing all substantive content.

\subsection{Case 5: Severe Style-Content Mismatch}

\textbf{Correct Response}: 
\begin{quote}
\texttt{"I want to see vast nature where water and greenery coexist, like a valley. I'd also love to see animals living freely on a savanna."}
\end{quote}

\textbf{Predicted Response}: 
\begin{quote}
\texttt{"I would love to shoot roll clouds, rainbows, and wild horses."}
\end{quote}

The model correctly identifies the photography context and maintains enthusiastic tone (M5: 5/5), but provides a much more concise list format versus the user's descriptive, expansive style. Moderate WMD (M1: 0.379), weak content match (M3: 2/5), and poor length match (M4: 0.481) illustrate how the model can match surface-level style while missing deeper personal expression patterns.

\subsection{Key Insights from Cases}

The case studies reveal several critical patterns: high style scores can be misleading, with perfect style matching (M5=5) often occurring even when responses are substantively wrong; length similarity emerges as a strong indicator, with M4 < 0.3 almost always indicating failure to capture user intent; history similarity consistently remains low (M6 rarely exceeds 0.5), indicating systemic failure to maintain user consistency; best performance requires alignment between user's natural style and model's analytical tendencies; and semantic metrics alone prove insufficient, as low WMD doesn't guarantee appropriate responses in conversational context.

\end{document}